\documentclass[10pt,journal,compsoc]{IEEEtran}
\ifCLASSOPTIONcompsoc
  \usepackage[nocompress]{cite}
\else
  \usepackage{cite}
\fi
\ifCLASSINFOpdf
\else
\fi
\usepackage{times}
\usepackage{latexsym}

\usepackage{microtype}

\usepackage{url}  
\usepackage{graphicx}  
\usepackage{multirow}
\usepackage{amsmath}
\usepackage{bm}
\usepackage{booktabs}
\usepackage[numbers]{natbib}
\usepackage[switch]{lineno}

\begin{document}
%
\title{Emotion-Regularized Conditional Variational Autoencoder for Emotional Response Generation}
%
%
%
%

\author{Yu-Ping~Ruan,
and Zhen-Hua~Ling,~\IEEEmembership{Senior Member,~IEEE,}

\IEEEcompsocitemizethanks{
\IEEEcompsocthanksitem Y.-P. Ruan is with the National University of Defense Technology, Hefei, 230031, China (ypruan@mail.ustc.edu.cn).
This work was done while he was studying at the University of Science and Technology of China.
\IEEEcompsocthanksitem Z.-H. Ling is with the National Engineering Laboratory of Speech and Language Information Processing,
University of Science and Technology of China, Hefei, 230027, China (e-mail: zhling@ustc.edu.cn).
}
\thanks{Manuscript received October 25, 2020; revised January 28, 2021.}}

%
%

\markboth{Journal of \LaTeX\ Class Files,~Vol.~14, No.~8, August~2015}%
{Shell \MakeLowercase{\textit{et al.}}: Bare Demo of IEEEtran.cls for Computer Society Journals}
%



\IEEEtitleabstractindextext{%
\begin{abstract}
This paper presents an emotion-regularized conditional variational autoencoder (Emo-CVAE) model for generating emotional conversation responses.
In conventional CVAE-based emotional response generation, emotion labels are simply used as additional conditions in prior, posterior and decoder networks.
Considering that emotion styles are naturally entangled with semantic contents in the language space, the Emo-CVAE model utilizes emotion labels to regularize the CVAE latent space by introducing an extra emotion prediction network.
In the training stage, the estimated latent variables are required to predict the emotion labels and token sequences of the input responses simultaneously. Experimental results show that our Emo-CVAE model can learn a more informative and structured latent space than a conventional CVAE model and output responses with better content and emotion performance than baseline CVAE and sequence-to-sequence (Seq2Seq) models.
\end{abstract}

}

\maketitle

\IEEEdisplaynontitleabstractindextext

%
\IEEEpeerreviewmaketitle

\IEEEraisesectionheading{\section{Introduction}\label{sec:introduction}}

%
%
%
%

\IEEEPARstart{T}{here} has been growing interest in building conversation agents that can chat with humans directly in natural language. Because emotion is a vital part of human intelligence \cite{mayer1993intelligence}, generating responses with specific emotions is one of the key steps towards making conversation agents more human-like. Existing studies have provided much evidence that systems capable of expressing emotions in their responses can improve user satisfaction \cite{partala2004effects,prendinger2005empathic}.

Traditionally, dialog systems depend on manual efforts to establish rules, design features, or build models with particular learning algorithms \cite{williams2007partially,schatzmann2006survey,young2013pomdp}. In early studies on emotional dialog systems, manually designed rules were used to deliberately select the desired ``emotional'' responses from a conversation corpus \cite{polzin2000emotion,skowron2010affect}; however, this approach is difficult or impossible to extend to unseen situations and large-scale datasets.

Benefiting from the development of deep learning techniques, neural-network-based models,
e.g., sequence-to-sequence (Seq2Seq) models \cite{vinyals2015neural,serban2016building,shang2015neural,li2016diversity} and variational generative models \cite{serban2017hierarchical,zhao2017learning,cao2017latent,ruan2019condition},
have achieved impressive results in training end-to-end generative chatting machines.
Several studies based on Seq2Seq and CVAE models
have been devoted to generating emotional responses \cite{zhou2018emotional,song2019generating,zhou2018mojitalk};
these studies all adopted RNNs as building blocks  and did not take advantage of pretrained language models.
Many recent studies have proven that language models pretrained on large-scale corpora can achieve state-of-the-art results on both language understanding and generation tasks \cite{radford2018improving,devlin2018bert,yang2019xlnet,dong2019unified}.
Thus, it is also worthwhile to utilize pretrained language models as building blocks for emotional response generation.

In this paper, we propose an emotion-regularized conditional variational autoencoder (Emo-CVAE) model for emotional response generation. The Emo-CVAE model uses BERT \cite{devlin2018bert}, one of the most popular pretrained language models, as the basis for its encoder and decoder. Different from the conventional CVAE model adopted in a previous study \cite{zhou2018mojitalk}, which simply used the emotion label $e$ as an extra input condition, the Emo-CVAE model utilizes the emotion label $e$ to regularize the latent distribution. Specifically, the estimated latent variable $\mathbf{z}$ in Emo-CVAE is used to recover the emotion label and token sequence of the input response $y$ simultaneously, considering that emotional style and content are naturally entangled in the language space. In experiments on the generation of short text conversations, we have found that the Emo-CVAE model can learn a more informative and structured latent distribution than the conventional CVAE model and output responses with better content and emotion performance than CVAE and Seq2Seq baselines.

\begin{figure*}[!ht]
	\centering
	\includegraphics[width=5.5in]{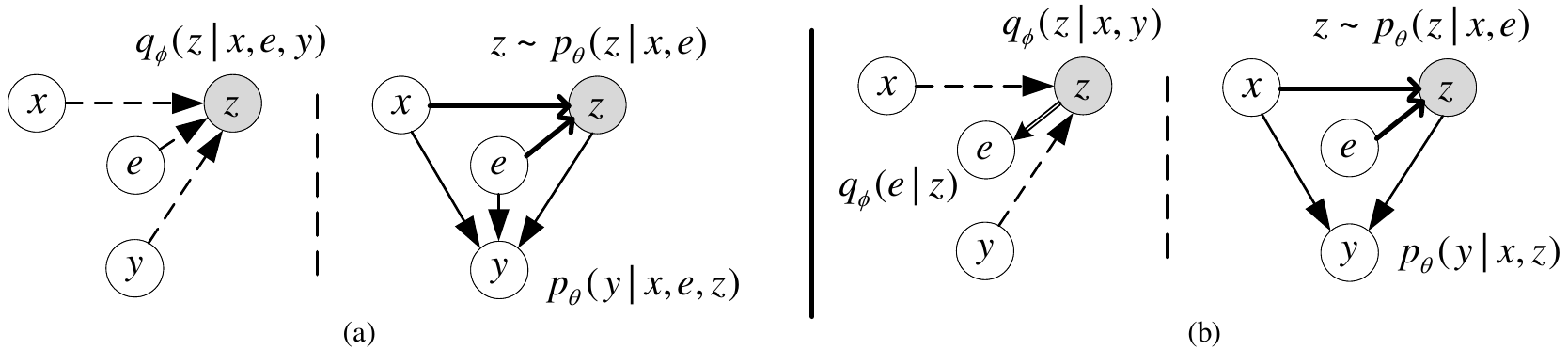}
\vspace{-4mm}
\caption{Graphical models of (a) a CVAE and (b) an Emo-CVAE, where $x$, $y$, and $e$ represent the input post (i.e., the conversation context), the response, and the emotion label of $y$, respectively. In each subgraph, the left part shows the process of the posterior estimation of the latent variable $\mathbf{z}$ during the training stage, and the right part shows the process of generating $y$ during the testing stage.
}\label{fig:graphical}
\end{figure*}

\section{Related Work} \label{sec:related_work}
Recently, several neural-network-based models have been proposed for emotional response generation. \citet{zhou2018emotional} proposed an emotional chatting machine (ECM), in which an external emotion vocabulary and an internal emotion state memory were introduced for better emotional expressiveness. \citet{song2019generating} proposed an emotional dialog system (EmoDS), in which a sentence-level emotion classifier and an external emotion vocabulary were used to guide the response generation process. Both the ECM and EmoDS models were based on the Seq2Seq framework.
Considering that CVAE models can perform better than Seq2Seq models for the generation of diverse responses \cite{serban2017hierarchical,zhao2017learning,ruan2019condition},
a CVAE-based emotional response generation method has also been proposed for Twitter conversations \cite{zhou2018mojitalk}.

There are still some deficiencies with existing studies on neural-network-based emotional response generation.
First, the ECM \cite{zhou2018emotional}, EmoDS \cite{song2019generating}, and CVAE \cite{zhou2018mojitalk} models mentioned above all employ RNNs as building blocks and do not take advantage of pretrained language models. 
Second, the conventional CVAE model adopted in a previous study \cite{zhou2018mojitalk} simply utilized the emotion label $e$ as an extra input condition, ignoring the interrelation between contents and emotional styles in the language space of responses.

To address these shortcomings, our Emo-CVAE model uses BERT \cite{devlin2018bert} as the basis for its encoder and decoder.
In addition, considering the difficulty of disentangling the stylistic properties of a sentence from its semantic content \cite{elazar2018adversarial,lample2018multiple},
the Emo-CVAE model constructs an entangled latent space regularized by emotional attributes,
which has proven to be beneficial for learning a more informative and structured latent distribution in our experiments. 


\begin{figure*}[!ht]
	\centering
	\includegraphics[width=5.3in]{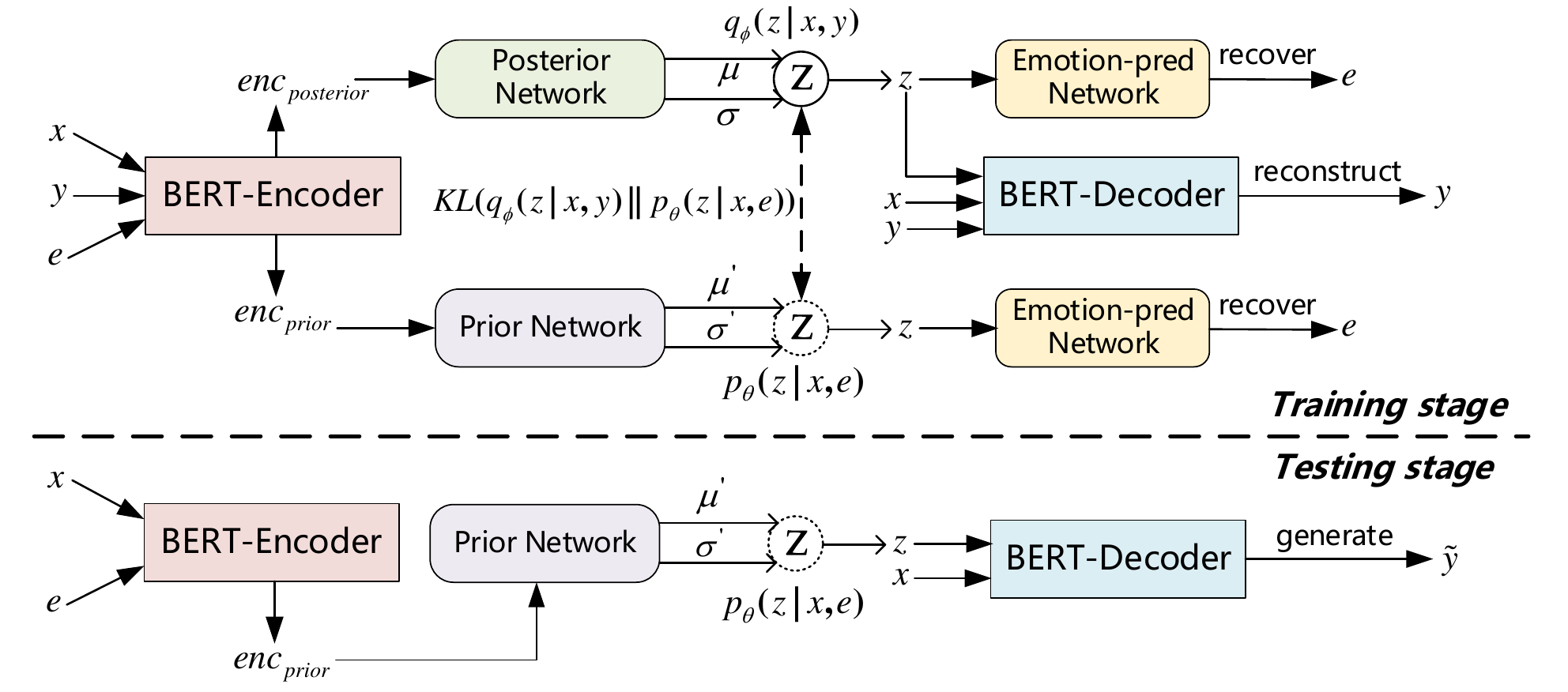}
\vspace{-4mm}
\caption{The architecture of the Emo-CVAE model implemented in this paper.}\label{fig:model}
\end{figure*}

\section{Methodology}
\subsection{From CVAE to Emo-CVAE}
Figure \ref{fig:graphical} shows the directed graphical models of the conventional CVAE approach and our proposed Emo-CVAE approach.
As Figure \ref{fig:graphical}(a) shows, in the CVAE model applied for emotional response generation in \cite{zhou2018mojitalk}, the input post $x$ and the emotion label $e$ are jointly used as the input conditions for estimating the latent variable $\mathbf{z}$ and generating the response $y$.
Specifically, a CVAE is composed of a prior network $p_\theta(\mathbf{z}|x,e)$, a posterior network $q_\phi(\mathbf{z}|x,e,y)$, and a decoder network $p_\theta(y|x,e,\mathbf{z})$. Both $p_\theta(\mathbf{z}|x,e)$ and $q_\phi(\mathbf{z}|x,e,y)$ are multivariate diagonal Gaussian distributions. Its model parameters can be efficiently trained within the stochastic gradient variational Bayes (SGVB) framework \cite{KingmaW13} by maximizing the lower bound on the conditional log likelihood $\log p(y|x,e)$, as follows:
\begin{equation}\label{eq:cvae_loss}
\begin{aligned}
\mathcal{L}&(\theta,\phi;x,e,y)=-KL(q_\phi(\mathbf{z}|x,e,y)||p_\theta(\mathbf{z}|x,e))\\
&+\mathbf{E}_{q_\phi(\mathbf{z}|x,e,y)}[\log p_\theta(y|x,e,\mathbf{z})]\le\log p(y|x,e).
\end{aligned}
\end{equation}

One deficiency of CVAE-based emotional response generation is that the posterior network $q_\phi(\mathbf{z}|x,e,y)$ is conditioned on both $e$ and $y$, in conflict with the fact that the emotion label $e$ and the semantic content of the response $y$ are entangled in $y$ itself (i.e., $e$ can be inferred from $y$).
Furthermore, $e$ is also used as a condition for decoding $y$, which causes the latent distribution in the CVAE to be independent of the input condition $e$.

Therefore, in the proposed Emo-CVAE model, the input condition $e$ is eliminated from both the posterior and decoder networks.
Instead, both the emotion label $e$ and the response $y$ are required to be recovered from the latent variable $\mathbf{z}$ simultaneously; to this end, an entangled latent distribution is adopted.
Specifically, as shown in Figure \ref{fig:graphical}(b), an Emo-CVAE includes a prior network $p_\theta(\mathbf{z}|x,e)$, a posterior network $q_\phi(\mathbf{z}|x,y)$, and a decoder network $p_\theta(y|x,\mathbf{z})$.
In addition, an emotion prediction network $q_\phi(e|\mathbf{z})$ is introduced to recover the emotion label $e$ from the sampled latent variable $\mathbf{z}$ and to regularize the CVAE latent space.

The model parameters of an Emo-CVAE are estimated by maximizing $\mathcal{L} = \mathcal{L}_{sgvb} + \mathcal{L}_{emo}$.
Following the training strategy for CVAEs, $\mathcal{L}_{sgvb}$ denotes 
the lower bound on the conditional log likelihood $\log p(y|x,e)$, as follows:
\begin{equation}\label{eq:emocvae_loss1}
\begin{aligned}
\mathcal{L}&_{sgvb}(\theta,\phi;x,e,y)=-KL(q_\phi(\mathbf{z}|x,y)||p_\theta(\mathbf{z}|x,e))\\
&+\mathbf{E}_{q_\phi(\mathbf{z}|x,y)}[\log p_\theta(y|x,\mathbf{z})]\le\log p(y|x,e).
\end{aligned}
\end{equation}
$\mathcal{L}_{emo}$ is defined as
\begin{equation}\label{eq:emocvae_loss2}
\begin{aligned}
\mathcal{L}_{emo}&(\theta,\phi;x,e,y)=\mathbf{E}_{q_\phi(\mathbf{z}|x,y)}[\log q_\phi(e|\mathbf{z})]\\
&+\mathbf{E}_{p_\theta(\mathbf{z}|x,e)}[\log q_\phi(e|\mathbf{z})]
\end{aligned}
\end{equation}
and acts as an emotion regularizer on the latent variables sampled using either the posterior or prior network.

\subsection{Model Implementation}
The architecture of the Emo-CVAE model implemented in this paper is shown in Figure \ref{fig:model}, and detailed introductions to the model components are given in this subsection.

\subsubsection{BERT-Encoder} 
\begin{figure}[!tb]
	\centering
	\includegraphics[width=3.5in]{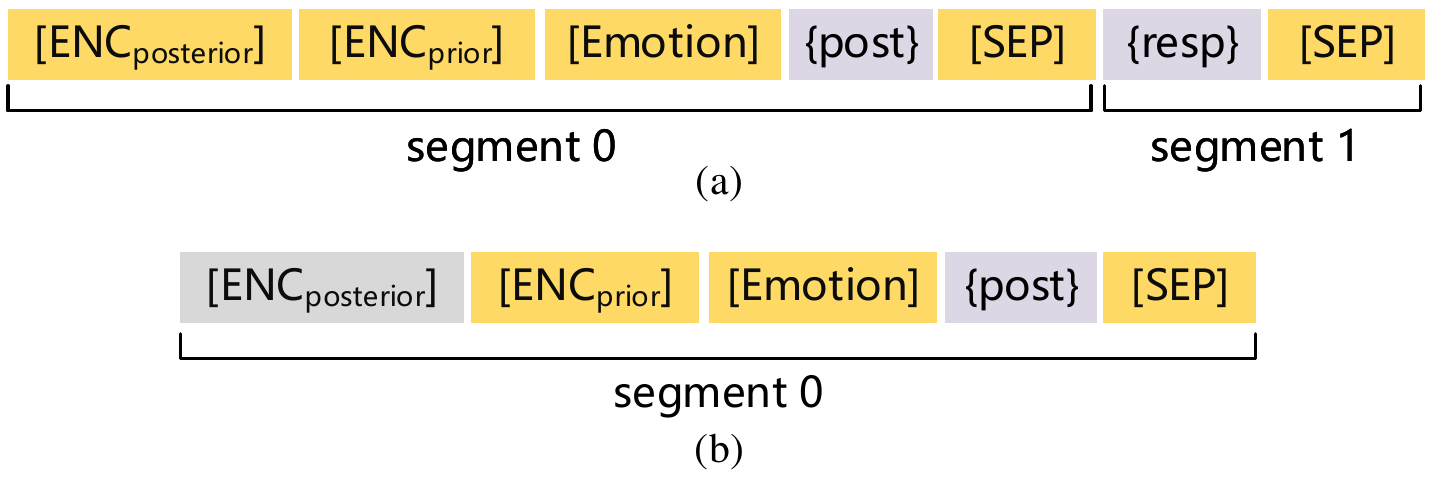}
\vspace{-8mm}
\caption{The input token sequences for the \emph{BERT-Encoder} in (a) the training stage and (b) the testing stage.}\label{fig:enc_inputs}
\end{figure}

The \emph{BERT-Encoder} is built on the basis of the pretrained BERT model \cite{devlin2018bert}. Its inputs in the training stage include the post $x$, the response $y$, and the emotion label $e$ of $y$. The input token sequence is shown in Figure \ref{fig:enc_inputs}(a), in which ``[Emotion]'', ``[ENC$_{posterior}$]'' and ``[ENC$_{prior}$]'' are tokens standing for the emotion label $e$, the encoding results for the posterior distribution, and the encoding results for the prior distribution, respectively. \{post\} and \{resp\} represent the token sequences of $x$ and $y$, respectively.
The top-layer encoding vectors of  ``[ENC$_{posterior}$]'' and ``[ENC$_{prior}$]'' correspond to the vectors $\mathbf{enc}_{posterior}$ and $\mathbf{enc}_{prior}$ shown in Figure \ref{fig:model},
which are further sent to the \emph{Posterior Network} and the \emph{Prior Network} to produce the distributions $q_\phi(\mathbf{z}|x,y)$ and $p_\theta(\mathbf{z}|x,e)$, respectively.
In the testing stage, the input token sequence for the \emph{BERT-Encoder} is as shown in Figure \ref{fig:enc_inputs}(b), and only $\mathbf{enc}_{prior}$ is output.

\begin{table}[!tb]
	\centering
	\renewcommand{\arraystretch}{1.3}
	\caption{The attention relationships among the tokens in the \emph{BERT-Encoder}. ``[SEP]''(0) and ``[SEP]''(1) represent the ``[SEP]''s in segments 0 and 1, respectively.}
\vspace{-3mm}
	\begin{tabular}{l|l}
		\hline
		& \bf{Attended Tokens} \\
         \hline
        ``[ENC$_{posterior}$]'' & ``[ENC$_{posterior}$]'', \{post\}, \{resp\} \\
        \hline
        ``[ENC$_{prior}$]'' & ``[ENC$_{prior}$]'', ``[Emotion]'', \{post\} \\
        \hline
        ``[Emotion]'' & ``[Emotion]'' \\
        \hline
        tokens in \{post\} &  \{post\}, ``[SEP]''(0) \\
        \hline
        ``[SEP]''(0) & ``[SEP]''(0), \{post\} \\
        \hline
        tokens in \{resp\} &  \{resp\}, ``[SEP]''(1) \\
        \hline
        ``[SEP]''(1) & ``[SEP]''(1), \{resp\} \\
        \hline
	\end{tabular}
	\label{tab:enc_att}
\end{table}

As shown in Table \ref{tab:enc_att}, the attention relationships among the tokens in the \emph{BERT-Encoder} are designed in accordance with the dependencies among the variables in $q_\phi(\mathbf{z}|x,y)$ and $p_\theta(\mathbf{z}|x,e)$.
For example, the ``[ENC$_{posterior}$]'' token attends to ``[ENC$_{posterior}$]'', \{post\}, and \{resp\} because $\mathbf{enc}_{posterior}$ is used to generate the distribution $q_\phi(\mathbf{z}|x,y)$.

\subsubsection{BERT-Decoder} 
\begin{figure}[!tb]
	\centering
	\includegraphics[width=2.3in]{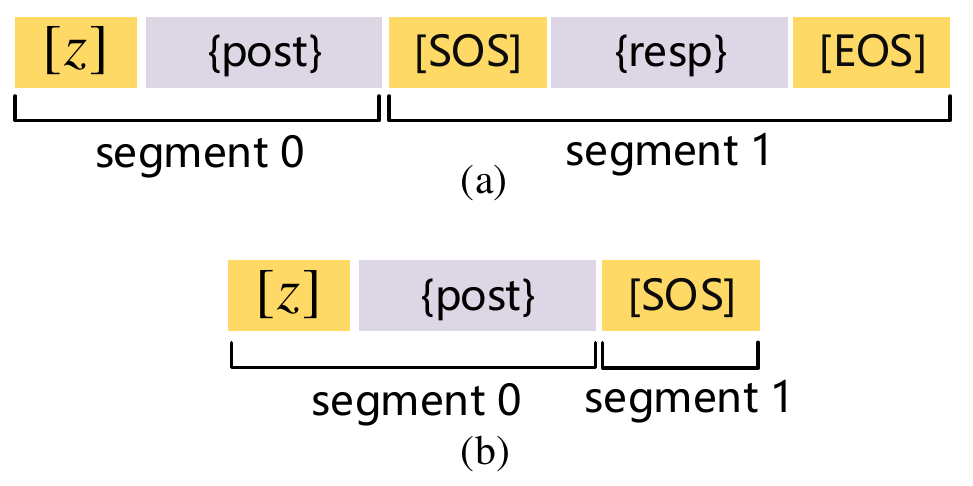}
\vspace{-4mm}
\caption{(a) The input token sequence of the \emph{BERT-Decoder} in the training stage and (b) the initial input token sequence of the \emph{BERT-Decoder} in the testing stage.}\label{fig:dec_inputs}
\end{figure}

The \emph{BERT-Decoder} is also based on the pretrained BERT model \cite{devlin2018bert}. Its input token sequence in the training stage is shown in Figure \ref{fig:dec_inputs}(a), in which ``[z]'', ``[SOS]'', and ``[EOS]'' stand for the sampled latent vector, the start flag token for a response, and the end flag token for a response, respectively. \{post\} and \{resp\} represent the token sequences of $x$ and $y$, respectively.
Figure \ref{fig:dec_inputs}(b) presents the initial input token sequence in the testing stage when generating a response.
The attention relationships among the tokens in the \emph{BERT-Decoder}, as shown in Table \ref{tab:dec_att},
are designed to generate a response from $p_\theta(y|x,\mathbf{z})$ in a token-by-token manner.
Thus, each token in \{resp\} attends to both the tokens in \{post\} and its historical tokens in \{resp\}.
The teacher-forcing mode is adopted when training the \emph{BERT-Decoder}.
In the testing stage, once a token in \{resp\} has been predicted, it is appended to the end of the token sequence and is attended to by subsequent tokens.
This process is repeated until the ``[EOS]'' token is predicted.

\subsubsection{Other modules} The \emph{Posterior Network} is a linear output layer that accepts $\mathbf{enc}_{posterior}$ as input and outputs $\mathbf{\bm{\mu}}$ and $\log(\mathbf{\bm{\sigma}}^2)$ for the posterior distribution $q_\phi(\mathbf{z}|x,y)=\mathcal{N}\mathbf{(\bm{\mu}, {\bm{\sigma}}^2I)}$. The \emph{Prior Network} is structurally identical to the \emph{Posterior Network}; it accepts $\mathbf{enc}_{prior}$ as input and makes predictions for the prior distribution $p_\theta(\mathbf{z}|x,e)$. The \emph{Emotion-pred Network} is a multilayer perceptron (MLP); in our implementation, it has a hidden layer with \emph{tanh} activation and a linear output layer.

\begin{table}[!tb]
	\centering
	\renewcommand{\arraystretch}{1.3}
	\caption{The attention relationships among the tokens in the \emph{BERT-Decoder}.}
\vspace{-3mm}	
    \begin{tabular}{l|p{4.0cm}}
		\hline
		& \bf{Attended Tokens} \\
         \hline
        ``[z]'' & ``[z]'', \{post\} \\
        \hline
        tokens in \{post\} & ``[z]'', \{post\} \\
        \hline
        ``[SOS]'' & ``[z]'', \{post\}, ``[SOS]'' \\
        \hline
        tokens in \{resp\} & all except future tokens in \{resp\} and ``[EOS]''\\
        \hline
        ``[EOS]'' & all tokens \\
        \hline
	\end{tabular}
	\label{tab:dec_att}
\end{table}

\subsection{Reranking Multiple Responses} \label{sec:rerank}
To help model output responses with better content relevance and emotional expressiveness, multiple responses to each post are first generated in the testing stage and are then reranked. 
Specifically, for variational generative models, we first generate multiple samples of $\mathbf{z}$, and then, a beam search is adopted for each $\mathbf{z}$ sample to obtain the best result. For Seq2Seq models, we first use the top-N candidates for the head word as seed words. Then, we continue to generate a response from each seed word through beam search decoding.

The score used for reranking the multiple responses to each post is defined as
\begin{equation}\label{eq:rank}
\begin{aligned}
score = {score}_{rele} + \lambda * {score}_{emo},
\end{aligned}
\end{equation}
where $\lambda$ is a constant weight and ${score}_{emo}$ and ${score}_{rele}$ are explained as follows.

(1) ${score}_{emo}$: The emotional expressiveness score is determined by a trained emotion classifier (see Section \ref{sec:dataset}). If the predicted emotion label of the generated response is consistent with the corresponding input emotion label, ${score}_{emo}=1$. Otherwise, ${score}_{emo}=0$.

(2) ${score}_{rele}$: The topic relevance score is provided by a topic coherence discrimination (TCD) model built by fine-tuning BERT \cite{devlin2018bert}. Specifically, we concatenate the post and the response to serve as the input token sequence for BERT and define the objective to judge whether the response is a valid response to the given post. The negative samples for training the TCD model are collected by randomly shuffling the mapping between the posts and responses. For a generated response $\tilde{y}$, we have ${score}_{rele} = p_{TCD}(true|x,\tilde{y})$.

\section{Experimental Setup}
\subsection{Dataset} \label{sec:dataset}
In our experiments, we adopted the short text conversation (STC) dataset from NTCIR-12 \cite{shang2016overview}, which was crawled from the Chinese-language platform Sina Weibo\footnote{\url{https://weibo.com/}}, one of the most popular social media platforms in China, and contains $1,800,000$ post-response pairs\footnote{This dataset was originally prepared for retrieval models and thus has no standard division for generative models. Here, we filtered the post-response pairs in the raw STC dataset in accordance with the word frequencies to build our dataset.}. We randomly split the data into $1,708,415$, $73,120$, and $18,465$ pairs to build the training, development and test sets, respectively.
There were no overlapping posts among these three sets.

Following previous studies \cite{zhou2018emotional,song2019generating}, we first trained an emotion classifier on the NLPCC dataset\footnote{\url{http://http://tcci.ccf.org.cn/nlpcc.php}} and then annotated the responses in the STC dataset using this classifier. Specifically, the NLPCC dataset we used contains $61,483$ sentences collected from Weibo, which were manually annotated into 8 emotion categories, i.e., liking, disgust, happiness, sadness, anger, surprise, fear, and other. The NLPCC dataset was partitioned into training, development, and test sets at a ratio of 8:1:1. The emotion classifier was built by fine-tuning BERT and ultimately achieved an accuracy of 68\% on the NLPCC test set.
The percentages of responses with different emotion annotations in the STC training set are shown in Table \ref{tab:emo_ratio}.

\begin{table}[!tb]
	\centering
	\renewcommand{\arraystretch}{1.3}
	\caption{The percentages of responses with different emotion annotations in the STC training set.}
	\vspace{-3mm}
    \begin{tabular}{cccc}
		\hline
		{liking} & {disgust} & {happiness} & {sadness}\\
         24.30\% & 9.89\% & 8.20\% & 6.90\% \\
         \hline
        {anger} & {surprise} & {fear} & {other}\\
        2.34\% & 5.08\% & 1.18\% & 42.11\% \\
        \hline
	\end{tabular}
	\label{tab:emo_ratio}
\end{table}

\subsection{Baselines}
In our experiments, we compared the Emo-CVAE model with four baselines, i.e., EmoDS, ECM, Seq2Seq, and CVAE models.

$\bullet$ \textbf{EmoDS} We implemented the EmoDS model \cite{song2019generating}, which is based on the Seq2Seq framework, using LSTM-RNNs as building blocks.

$\bullet$ \textbf{ECM} We implemented the ECM model \cite{zhou2018emotional}, which is also based on the Seq2Seq framework, using GRU-RNNs as building blocks.

$\bullet$ \textbf{Seq2Seq} To ensure better fairness in the comparisons between the Seq2Seq models and our Emo-CVAE model, we also implemented a Seq2Seq model for emotional response generation based on BERT. Specifically, the Seq2Seq model was structurally identical to the \emph{BERT-Decoder} in the Emo-CVAE model, but the input token ``[z]'' in the Emo-CVAE model was replaced with the token ``[Emotion]'' for representing emotion labels.

$\bullet$ \textbf{CVAE} The CVAE model used in a previous work \cite{zhou2018mojitalk} was based on GRU-RNNs. To support a fairer comparison between the CVAE and Emo-CVAE approaches, we implemented a CVAE model based on BERT. Specifically, the CVAE implementation was almost the same as the Emo-CVAE implementation, with three differences. First, a token ``[Emotion]'' for emotion labels was concatenated with the head of the input token sequence in the \emph{BERT-Decoder}. Second, the attention flow settings in the \emph{BERT-Encoder} and \emph{BERT-Decoder} were revised to satisfy the requirements of the prior network $p_\theta(\mathbf{z}|x,e)$, the posterior network $q_\phi(\mathbf{z}|x,e,y)$, and the decoder network $p_\theta(y|x,e,\mathbf{z})$. Third, there was no \emph{Emotion-pred Network} in the CVAE model, and the $\mathcal{L}_{emo}$ term in the Emo-CVAE objective function was eliminated accordingly.

\begin{figure}[!tb]
	\centering
	\includegraphics[width=3.0in]{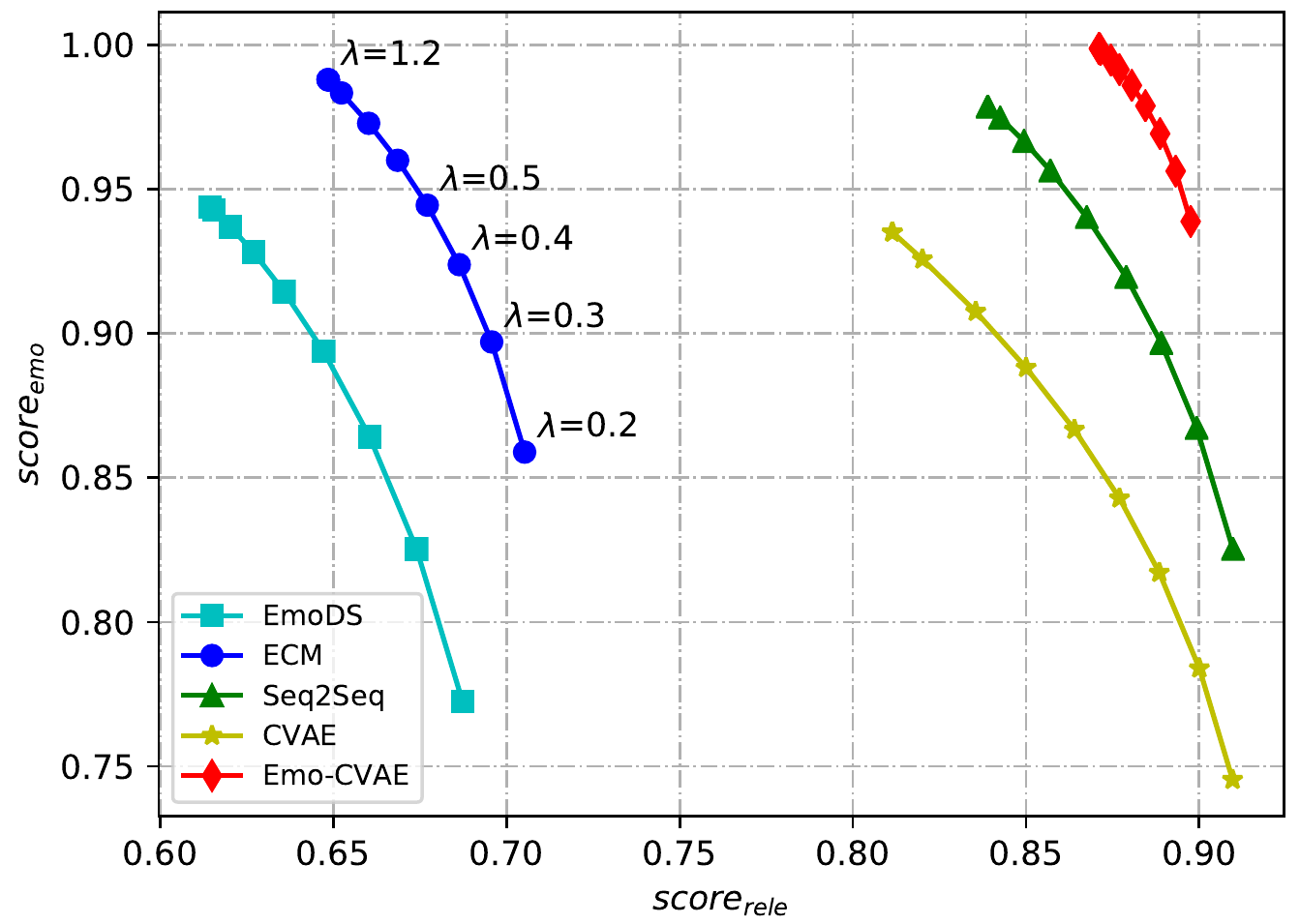}
\vspace{-4mm}
\caption{The average emotional expressiveness scores and topic relevance scores of the models' top-1 responses with different $\lambda$ values.}\label{fig:emo_rele}
\end{figure}

\begin{table*}[!tb]
	\centering
	\renewcommand{\arraystretch}{1.3}
	\caption{The objective evaluation results for the top-1 responses generated by the different models.}
\vspace{-2mm}
	\begin{tabular}{l|ccccc}
		\hline
		& {Emo Acc.~(\%)} & {Rele.} & {Distinct-1/2} & {Uniq.~(\%)} & {PPL on LM}\\
         \hline
        {EmoDS} & 89.37 & 0.6472 & 0.012/0.043 & 27.25 & 8.41 \\
        {ECM} & 94.44 & 0.6771 & 0.014/0.054 & 26.04 & \textbf{8.00} \\
        \hline
        {Seq2Seq} & 91.95 & 0.8789  & 0.070/0.311 & 80.40 & 11.02 \\
        {CVAE} & 84.29 & 0.8770 & 0.081/\textbf{0.421} & 97.15 & 14.88 \\
        {Emo-CVAE} & \textbf{97.88} & \textbf{0.8845} & \textbf{0.083}/0.419 & \textbf{97.16} & 14.15 \\
        \hline
	\end{tabular}
	\label{tab:obj_all}
\vspace{-2mm}
\end{table*}

\begin{table*}[!tb]
	\centering
	\renewcommand{\arraystretch}{1.3}
	\caption{The detailed emotion accuracies (\%) of the top-1 responses generated by the different models.}
	\vspace{-2mm}
    \begin{tabular}{l|c|cccccccc}
		\hline
		 & \emph{all}& {other} & {liking} & {disgust} & {happiness} & {sadness} & {anger} & {surprise} & {fear} \\
         \hline
        {EmoDS} & 89.37 & 98.27 & 98.27 & 68.78 & 99.42 & 92.40 & 64.86 & 98.39 & 94.59  \\
        {ECM} & 94.44 & 99.31 & 99.07 & 85.71 & \textbf{99.54} & 96.65 & 80.07 & \textbf{98.96} & 96.20 \\
        \hline
        {Seq2Seq} & 91.95 & 99.19 & \textbf{99.77} & 88.02 & 98.16 & 93.66 & 72.00 & 97.58 & 87.21 \\
        {CVAE} & 84.29 & 98.84 & 97.46 & 75.92 & 92.40 & 84.10 & 51.73 & 87.90 & 85.94 \\
        {Emo-CVAE} & \textbf{97.89} & \textbf{99.65} & 99.65 & \textbf{97.70} & 99.08 & \textbf{97.58} & \textbf{94.82} & 98.16 & \textbf{96.43} \\
        \hline
	\end{tabular}
	\label{tab:emo_acc}
\vspace{-2mm}
\end{table*}

\subsection{Parameter Settings}
The PyTorch implementation of BERT with the pretrained model file {\tt{bert-base-chinese}} provided by Google\footnote{\url{https://github.com/huggingface/pytorch-pretrained-BERT\#Fine-tuning-with-BERT-running-the-examples}}
was employed. The latent variable $\mathbf{z}$ in the CVAE and Emo-CVAE models had $768$ dimensions, and the hidden size of the \emph{Emotion-pred Network} was set to $768$. For the fine-tuning process, we mostly followed the default settings. Specifically, the learning rate was $2e$-$05$, and the batch size was set to $128$. When training the CVAE and Emo-CVAE models, the KL annealing strategy was adopted to address the issue of latent variable vanishing. The model parameters were pretrained without optimizing the KL divergence term.
Additionally, we adopted a training strategy of optimizing the KLD loss term every 15 steps but optimizing the reconstruction nonnegative log likelihood (NLL) loss term every step.
As described in Section \ref{sec:rerank}, we generated multiple responses to each post. Specifically, 5 responses were generated for each unique post in the test set concerning different emotional inputs, and the beam search size was 5.

\section{Results}
\subsection{Objective Evaluations}
The objective evaluation metrics used in our experiments included the following:
\begin{enumerate}[\IEEEsetlabelwidth{12)}]
\item {Emotional accuracy} (\emph{Emo Acc.}), which was determined by the trained emotion classifier introduced in Section \ref{sec:dataset} (i.e., the average of ${score}_{emo}$ on a test set) and evaluates the models' ability to generate responses expressing specific emotions.
\item {Relevance score} (\emph{Rele.}), which was determined by the trained TCD model introduced in Section \ref{sec:rerank} (i.e., the average of ${score}_{rele}$ on a test set) and evaluates whether the generated responses are relevant to the topic of the input post.
\item {Diversity performance scores} (\emph{distinct-1}, \emph{distinct-2}, and \emph{Uniq.}), which represent the percentages of distinct unigrams, bigrams \cite{li2016diversity}, and sentences \cite{ruan2019condition} in the generated responses. The \emph{distinct-1/2} and \emph{Uniq.} scores evaluate the generated responses at the n-gram level and the sentence level, respectively.
\item {Fluency performance} (\emph{PPL on LM}), which represents the perplexity of the generated responses evaluated using a BERT language model trained on the same STC dataset.
It should be noted that better diversity performance generally leads to higher LM perplexity for a specific model, which implies worse performance in terms of fluency.
\end{enumerate}

\subsubsection{The effects of different $\lambda$ weights}
As introduced in Section \ref{sec:rerank}, multiple responses were generated and reranked for each post in our experiments. To investigate the influence of the weight $\lambda$ in Equation (\ref{eq:rank}) on the reranking results of the different models, we varied the value of $\lambda$ in increments of $0.1$ in the range of $[0.2, 1.2]$, and the average reranking scores of the models' top-1 responses are presented in Figure \ref{fig:emo_rele}.
There were trade-offs between ${score}_{emo}$ and ${score}_{rele}$ for all models, and a larger $\lambda$ always led to higher accuracy of emotional expression.
There was a large performance gap between the RNN-based models (i.e., EmoDS and ECM) and the BERT-based models (i.e., Seq2Seq, CVAE, and Emo-CVAE), which indicates the effectiveness of using the pretrained BERT model \cite{devlin2018bert} for emotional response generation. Compared with the baselines, the Emo-CVAE model obtained the highest overall reranking scores. 

To simplify further objective and subjective evaluations, $\lambda$ was fixed to $0.5$ in the following experiments.
\subsubsection{Overall performance}
Table \ref{tab:obj_all} presents the objective evaluation results for the top-1 responses generated by the different models.

Table \ref{tab:obj_all} shows that the performance of the RNN-based models, i.e., EmoDS and ECM, was obviously worse than that of the BERT-based models in terms of the relevance and diversity metrics. The variational generative models, i.e., CVAE and Emo-CVAE, achieved better performance in generating diverse responses than the models based on the sequence-to-sequence framework,
which is consistent with the findings of previous studies \cite{ruan2019condition}.
Specifically, our Emo-CVAE model outperformed the CVAE model and achieved the best performance in terms of both emotional accuracy and relevance. Regarding LM perplexity, both the Emo-CVAE and CVAE models performed worse than the other models. This is reasonable since better diversity generally leads to higher LM perplexity. Compared with the CVAE model,
our Emo-CVAE model achieved almost the same diversity performance but lower LM perplexity. 

Table \ref{tab:emo_acc} presents the detailed emotional accuracies of the top-1 responses generated by the different models. The advantages of Emo-CVAE over other models mainly lie in its expressiveness for the emotions of ``disgust'' and ``anger''.

\begin{table}[!tb]
	\centering
    \renewcommand{\arraystretch}{1.3}
    \caption{The proportions ($\%$) of the test samples conditioned on the emotional correctness of Emo-CVAE and the baselines when applied for response generation. Detailed explanations can be found in Section \ref{ref:subeva}.}
	\vspace{-2mm}
    \begin{tabular}{l|cccc}
		\hline
		{Model} & {TT} & {TF} & {FT} & {FF}\\
        \hline
         EmoDS & 87.48 & 10.35 & 1.69 & 0.48 \\
         ECM & 92.44 & 5.39 & 1.90 & 0.26 \\
         Seq2Seq & 90.31 & 7.59 & 1.56 & 0.54 \\
         CVAE & 82.73 & 15.17 & 1.36 & 0.74 \\
         \hline
	\end{tabular}
	\label{tab:sample_ratio}
\vspace{-2mm}
\end{table}

\begin{table}[!tb]
	\centering
    \renewcommand{\arraystretch}{1.3}
    \caption{Average preference scores (std.) (\%) based on \emph{emotion} and \emph{content}  when comparing Emo-CVAE with the baselines, where $p$ denotes the $p$-value of a t-test between two models.}
	\vspace{-2mm}
    \begin{tabular}{l|cccc}
		\toprule[1.0pt]
		 & \bf{Win} & \bf{Lose} & \bf{Tie} & \bf{$p$} \\
        \midrule[0.8pt]
        \multicolumn{5}{c}{{ Emo-CVAE vs. EmoDS (emotion)}} \\
        \hline
        (TT) & \textbf{34.4}(9.4) & 23.9(16.5) & 41.7(24.8) & $<0.01$\\
        (TF) & \textbf{63.9}(12.1) & 9.4(5.6) & 26.7(15.0) & $<0.001$ \\
        (FT) & 19.0(17.4) & \textbf{55.0}(7.1) & 26.0(19.8) & $<0.001$ \\
        (FF) & \textbf{33.0}(17.5) & 17.0(20.1) & 50.0(29.8) & $<0.01$ \\
        \hline
        \multicolumn{5}{c}{{ Emo-CVAE vs. ECM (emotion)}} \\
        \hline
        (TT) & 27.2(12.9) & 24.4(15.8) & 48.33(26.5) & $>\textbf{0.05}$ \\
        (TF) & \textbf{75.6}(10.4) & 3.9(4.6) & 20.6(13.2) & $<0.001$ \\
        (FT) & 11.0(10.7) & \textbf{50.0}(15.8) & 39.0(24.6) & $<0.001$ \\
        (FF) & \textbf{28.7}(21.1) & 15.0(17.9) & 56.2(34.0) & $<0.01$ \\
        \hline
        \multicolumn{5}{c}{{ Emo-CVAE vs. Seq2Seq (emotion)}} \\
        \hline
        (TT) & \textbf{33.9}(17.5) & 24.4(11.5) & 41.7(28.3) & $<0.05$ \\
        (TF) & \textbf{70.6}(10.3) & 9.4(10.1) & 20.0 (13.9) & $<0.001$ \\
        (FT) & 14.0(10.2) & \textbf{42.0}(17.2) & 44.0(26.9) & $<0.001$ \\
        (FF) & \textbf{29.0}(11.1) & 15.0(18.2) & 56.0(22.7) & $<0.01$ \\
        \hline
        \multicolumn{5}{c}{{ Emo-CVAE vs. CVAE (emotion)}} \\
        \hline
        (TT) & \textbf{23.9}(11.1) & 13.3(7.2) & 62.8(16.9) & $<0.001$ \\
        (TF) & \textbf{59.4}(19.8) & 6.7(3.8) & 33.9(23.3) & $<0.001$ \\
        (FT) & 12.8(8.0) & \textbf{42.8}(5.9) & 44.4(10.5) & $<0.001$\\
        (FF) & \textbf{32.2}(10.7) & 10.6(8.9) & 57.2(18.0) & $<0.001$\\
        \midrule[0.8pt]
        \multicolumn{5}{c}{{ Emo-CVAE vs. EmoDS (content)}} \\
        \hline
        (TT) & \textbf{42.8}(13.9) & 25.6(10.3) & 31.7(21.3) & $<0.001$ \\
        (TF) & \textbf{47.8}(11.8) & 19.4(9.3) & 32.8(18.7) & $<0.001$ \\
        (FT) & \textbf{52.0}(8.7) & 35.0(7.1) & 13.0(9.3) & $<0.01$ \\
        (FF) & 36.0(15.9) & 38.0(14.7) & 26.0(24.8) & $>\textbf{0.05}$ \\
        \hline
        \multicolumn{5}{c}{{ Emo-CVAE vs. ECM (content)}} \\
        \hline
        (TT) & \textbf{50.0}(14.3) & 22.2(11.4) & 27.8(19.5) & $<0.001$ \\
        (TF) & \textbf{42.8}(14.2) & 32.8(12.4) & 24.4(16.9) & $<0.05$ \\
        (FT) & 34.0(12.4) & 42.0(16.9) & 24.0(17.7) & $>\textbf{0.05}$ \\
        (FF) & 27.5(19.6) & \textbf{42.5}(10.8) & 30.0(23.2) & $<0.05$ \\
        \hline
        \multicolumn{5}{c}{{ Emo-CVAE vs. Seq2Seq (content)}} \\
        \hline
        (TT) & \textbf{32.8}(17.2) & 22.8(14.2) & 44.4(28.9) & $<0.05$ \\
        (TF) & \textbf{70.6}(10.3) & 9.4(10.1) & 20.0 (13.9) & $<0.001$ \\
        (FT) & 33.0(10.3) & \textbf{47.0}(14.0) & 20.0(16.4) & $<0.05$ \\
        (FF) & 32.0(11.2) & 33.0(13.3) & 35.0(22.6) & $>\textbf{0.05}$ \\
        \hline
        \multicolumn{5}{c}{{ Emo-CVAE vs. CVAE (content)}} \\
        \hline
        (TT) & \textbf{37.8}(12.7) & 26.1(8.0) & 36.1(12.8) & $<0.01$ \\
        (TF) & \textbf{47.2}(11.8) & 27.2(14.6) & 25.6(19.8) & $<0.001$ \\
        (FT) & 36.1(11.8) & 36.7(6.9) & 27.2(17.7) & $>\textbf{0.05}$\\
        (FF) & \textbf{30.6}(12.2) & 20.0(9.6) & 49.4(18.7) & $<0.01$\\
        \bottomrule[1.0pt]
	\end{tabular}
	\label{tab:preference}
\vspace{-5mm}
\end{table}

\subsection{Subjective Evaluations}
\label{ref:subeva}
Several groups of subjective ABX preference tests were conducted.
To make the human evaluations more efficient, a nonuniform random sampling strategy was adopted for selecting the test samples for human judgment. Table \ref{tab:sample_ratio} shows the proportions of the test samples conditioned on the emotional correctness of Emo-CVAE and the baselines when applied for response generation.
For example, the ``TF'' subset represents the samples for which the responses generated by Emo-CVAE had the correct emotional expression but the responses generated by a baseline model did not.
Then, 30 samples from each subset in Table \ref{tab:sample_ratio} were randomly selected, and six native Chinese speakers with rich Sina Weibo experience were recruited for an evaluation. For each test sample, a pair of responses generated by two of the models were presented in a random order.
Based on each of two subjective metrics, \emph{emotion} and \emph{content}, the evaluators were asked to judge which response in each pair was preferred or if there was no preference.
Here, \emph{emotion} refers to whether the emotional expression of a response agrees with the given emotion label, and \emph{content} refers to whether the response is relevant to the topic, informative, and fluent.
The $p$-value of a $t$-test was adopted to measure the significance of the difference between each pair of models.
Several significance levels were examined, including $p<0.05$, $p<0.01$, and $p<0.001$. A level of $p>0.05$ indicated that there was no significant difference between the two models.
The subjective evaluation results are presented in Table \ref{tab:preference}.

Table \ref{tab:preference} shows that the Emo-CVAE model significantly outperformed all baseline models in terms of both the \emph{emotion} and \emph{content} metrics for the ``TF'' subset. For the ``TT'' subset, our Emo-CVAE model also showed a significant advantage over the baseline models except when compared with the ECM model based on the \emph{emotion} metric.
For the ``FT'' and ``FF'' subsets, some baseline models outperformed the Emo-CVAE model. However, the proportions of responses belonging to the ``FT'' and ``FF'' subsets were much smaller than those belonging to the ``TT'' and ``TF'' subsets (see Table \ref{tab:sample_ratio}), which indicates that our Emo-CVAE model can generate responses with better overall content relevance and emotional expressiveness than the baselines.

\subsection{Ablation Studies}
\begin{table*}[!tb]
	\centering
    \renewcommand{\arraystretch}{1.3}
    \caption{The objective evaluation results for the top-1 responses generated by Emo-CVAE models with different emotion regularization settings.}
	\vspace{-3mm}
    \begin{tabular}{l|ccccc}
		\hline
		& {Emo Acc.~(\%)} & {Rele.} & {Distinct-1/2} & {Uniq.~(\%)} & {PPL on LM}\\
		\hline
		{Emo-CVAE} & 97.88 & 0.8845 & 0.083/0.419 & 97.16 & 14.15 \\
        {Emo-CVAE-M1} & 62.34 & 0.8970 & 0.084/0.426 & 97.51 & 14.87 \\
        {Emo-CVAE-M2} & 31.37 & 0.9128 & 0.082/0.441 & 99.11 & 15.18  \\
        \hline
	\end{tabular}
	\label{tab:ablation_0}
\end{table*}

\begin{table*}[!tb]
	\centering
    \renewcommand{\arraystretch}{1.3}
    \caption{The objective evaluation results for the top-1 responses generated by the different models in the ablation studies.}
	\vspace{-3mm}
    \begin{tabular}{l|ccccc}
		\hline
		& {Emo Acc.~(\%)} & {Rele.} & {Distinct-1/2} & {Uniq.~(\%)} & {PPL on LM}\\
        \hline
        {CVAE} & 84.29 & 0.8770 & 0.081/0.421 & 97.15 & 14.88 \\
        {CVAE-M1} & 30.85 & 0.9288 & 0.080/0.402 & 97.16 & 15.80  \\
        {CVAE-M2} & 86.06 & 0.8501 & 0.086/0.441 & 98.34 & 14.43  \\
        \hline
        {Emo-CVAE} & 97.88 & 0.8845 & 0.083/0.419 & 97.16 & 14.15 \\

        \hline
	\end{tabular}
	\label{tab:ablation}
\end{table*}

\subsubsection{Importance of emotion regularization}
To investigate the importance of emotion regularization in the Emo-CVAE model, we first removed the emotion regularization applied to the prior latent distribution, resulting in the ``Emo-CVAE-M1'' model. Then, the emotion regularization on the posterior latent distribution was also removed, yielding the ``Emo-CVAE-M2'' model. Table \ref{tab:ablation_0} presents the objective evaluation results for the top-1 responses generated by the Emo-CVAE models with different emotion regularization settings.

Table \ref{tab:ablation_0} shows that 
although the relevance score is slightly degraded with emotion regularization, emotion regularization on both the prior and posterior distributions is critically important for generating emotional responses of the expected emotional types.

\begin{figure*}[!tb]
	\centering
	\includegraphics[width=5.2in]{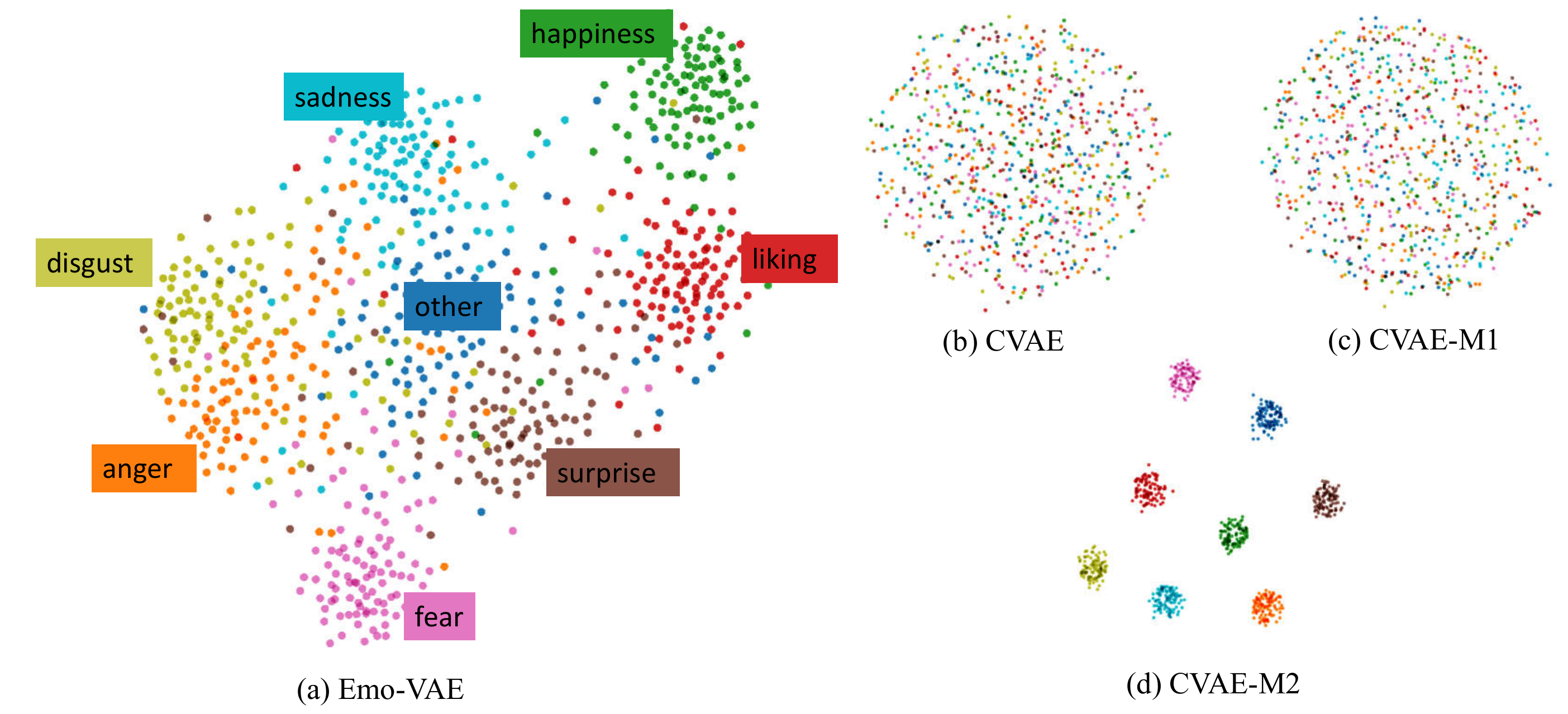}
\vspace{-3mm}
\caption{t-SNE visualization (in 2D space) of the posterior latent distributions produced by the (a) Emo-CVAE, (b) CVAE, (c) CVAE-M1, and (d) CVAE-M2 models.}\label{fig:latent_space}
\vspace{-2mm}
\end{figure*}

\subsubsection{Comparison between Emo-CVAE and CVAE}
Several ablation studies were conducted to further assess the advantages of the Emo-CVAE approach over the CVAE approach.
We gradually modified the structure of the CVAE model toward that of the Emo-CVAE model. 
First, we removed the input $e$ for the \emph{BERT-Decoder} in the CVAE model, resulting in the ``CVAE-M1'' model.
Then, the emotion regularization strategy was applied to the latent distributions in CVAE-M1 by introducing the emotion prediction network and the $\mathcal{L}_{emo}$ component for model training, yielding the ``CVAE-M2'' model. The only difference between CVAE-M2 and Emo-CVAE lies in their posterior distributions, i.e., $q_\phi(\mathbf{z}|x,e,y)$ and $q_\phi(\mathbf{z}|x,y)$, respectively.
Table \ref{tab:ablation} presents the objective evaluation results of the ablation studies.

Table \ref{tab:ablation} shows that the emotional accuracy of CVAE-M1 is dramatically degraded, indicating that the latent distribution learned in CVAE-M1 fails to accurately convey emotional information. The CVAE-M2 model achieves better emotional accuracy than either the CVAE-M1 model or the original CVAE model, demonstrating the effectiveness of our proposed emotion regularization strategy. However, CVAE-M2 still does not perform as well as the Emo-CVAE model.

To better analyze the above results, the latent distributions learned by the different models were visualized. Specifically, t-SNE \cite{maaten2008visualizing} was utilized to visualize the distributions of the posterior $\mathbf{z}$ samples (i.e., the posterior distributions).
As Figure \ref{fig:latent_space}(a) shows, the latent distribution in Emo-CVAE has a clear emotion-clustered structure, although there are blending trends among sample clusters corresponding to different emotions, which reflect the entanglement of emotion and content.
The $\mathbf{z}$ samples associated with similar emotions, such as ``happiness'' vs. ``liking'' and ``disgust'' vs. ``anger'', are distributed near each other.
On the other hand, the latent distributions $q_\phi(\mathbf{z}|x,e,y)$ learned by the CVAE and CVAE-M1 models appear independent of emotion, 
indicating that the emotional information from the input conditions $e$ and $y$ vanishes in the latent distributions $q_\phi(\mathbf{z}|x,e,y)$ of these models.
The CVAE-M2 model can learn a latent distribution with a strong emotion-clustered structure due to the help of emotion regularization. However, the distribution $q_\phi(\mathbf{z}|x,e,y)$ in CVAE-M2 is less informative than the distribution $q_\phi(\mathbf{z}|x,y)$ in Emo-CVAE.
The emotional information expressed by $q_\phi(\mathbf{z}|x,e,y)$ in CVAE-M2 can be simply copied from the input condition $e$, 
while to obtain $q_\phi(\mathbf{z}|x,y)$ in Emo-CVAE, it is necessary to infer the emotional information from the post $x$ and the response $y$.
Such a posterior distribution with inferred emotional information may be beneficial for learning prior distributions for response generation.

\section{Conclusion}
In this paper, we have proposed an emotion-regularized conditional variational autoencoder (Emo-CVAE) model for emotional conversation response generation. In this model, an entangled latent distribution is designed by forcing the estimated latent variable  to recover the emotion label and token sequence of a response simultaneously. In experiments on the generation of short text conversations, we have found that the Emo-CVAE model can learn a more informative and structured latent distribution than the conventional CVAE model and generate output responses with better content and emotional expressiveness than CVAE and Seq2Seq baselines.

\ifCLASSOPTIONcaptionsoff
\newpage
\fi



%



\bibliographystyle{IEEEtranN}
\bibliography{IEEEabrv,ref}

\begin{thebibliography}{28}
\providecommand{\natexlab}[1]{#1}
\providecommand{\url}[1]{#1}
\csname url@samestyle\endcsname
\providecommand{\newblock}{\relax}
\providecommand{\bibinfo}[2]{#2}
\providecommand{\BIBentrySTDinterwordspacing}{\spaceskip=0pt\relax}
\providecommand{\BIBentryALTinterwordstretchfactor}{4}
\providecommand{\BIBentryALTinterwordspacing}{\spaceskip=\fontdimen2\font plus
\BIBentryALTinterwordstretchfactor\fontdimen3\font minus
  \fontdimen4\font\relax}
\providecommand{\BIBforeignlanguage}[2]{{%
\expandafter\ifx\csname l@#1\endcsname\relax
\typeout{** WARNING: IEEEtranN.bst: No hyphenation pattern has been}%
\typeout{** loaded for the language `#1'. Using the pattern for}%
\typeout{** the default language instead.}%
\else
\language=\csname l@#1\endcsname
\fi
#2}}
\providecommand{\BIBdecl}{\relax}
\BIBdecl

\bibitem[MAYER and SALOVEY(1993)]{mayer1993intelligence}
J.~D. MAYER and P.~SALOVEY, ``The intelligence of emotional intelligence,''
  \emph{INTELLIGENCE}, vol.~17, pp. 433--442, 1993.

\bibitem[Partala and Surakka(2004)]{partala2004effects}
T.~Partala and V.~Surakka, ``The effects of affective interventions in
  human--computer interaction,'' \emph{Interacting with computers}, vol.~16,
  no.~2, pp. 295--309, 2004.

\bibitem[Prendinger and Ishizuka(2005)]{prendinger2005empathic}
H.~Prendinger and M.~Ishizuka, ``The empathic companion: A character-based
  interface that addresses users'affective states,'' \emph{Applied artificial
  intelligence}, vol.~19, no. 3-4, pp. 267--285, 2005.

\bibitem[Williams and Young(2007)]{williams2007partially}
J.~D. Williams and S.~Young, ``Partially observable {Markov} decision processes
  for spoken dialog systems,'' \emph{Computer Speech \& Language}, vol.~21,
  no.~2, pp. 393--422, 2007.

\bibitem[Schatzmann et~al.(2006)Schatzmann, Weilhammer, Stuttle, and
  Young]{schatzmann2006survey}
J.~Schatzmann, K.~Weilhammer, M.~Stuttle, and S.~Young, ``A survey of
  statistical user simulation techniques for reinforcement-learning of dialogue
  management strategies,'' \emph{The knowledge engineering review}, vol.~21,
  no.~2, pp. 97--126, 2006.

\bibitem[Young et~al.(2013)Young, Gasic, Thomson, and Williams]{young2013pomdp}
S.~Young, M.~Gasic, B.~Thomson, and J.~D. Williams, ``{POMDP}-based statistical
  spoken dialog systems: A review,'' \emph{Proceedings of the IEEE}, vol.~5,
  no. 101, pp. 1160--1179, 2013.

\bibitem[Polzin and Waibel(2000)]{polzin2000emotion}
T.~S. Polzin and A.~Waibel, ``Emotion-sensitive human-computer interfaces,'' in
  \emph{ISCA tutorial and research workshop (ITRW) on speech and emotion},
  2000.

\bibitem[Skowron(2010)]{skowron2010affect}
M.~Skowron, ``Affect listeners: Acquisition of affective states by means of
  conversational systems,'' in \emph{Development of Multimodal Interfaces:
  Active Listening and Synchrony}.\hskip 1em plus 0.5em minus 0.4em\relax
  Springer, 2010, pp. 169--181.

\bibitem[Vinyals and Le(2015)]{vinyals2015neural}
O.~Vinyals and Q.~Le, ``A neural conversational model,'' \emph{arXiv preprint
  arXiv:1506.05869}, 2015.

\bibitem[Serban et~al.(2016)Serban, Sordoni, Bengio, Courville, and
  Pineau]{serban2016building}
I.~V. Serban, A.~Sordoni, Y.~Bengio, A.~Courville, and J.~Pineau, ``Building
  end-to-end dialogue systems using generative hierarchical neural network
  models,'' in \emph{Thirtieth AAAI Conference on Artificial Intelligence},
  2016.

\bibitem[Shang et~al.(2015)Shang, Lu, and Li]{shang2015neural}
L.~Shang, Z.~Lu, and H.~Li, ``Neural responding machine for short-text
  conversation,'' in \emph{Proceedings of the 53rd Annual Meeting of the
  Association for Computational Linguistics and the 7th International Joint
  Conference on Natural Language Processing (Volume 1: Long Papers)}, 2015, pp.
  1577--1586.

\bibitem[Li et~al.(2016)Li, Galley, Brockett, Gao, and Dolan]{li2016diversity}
J.~Li, M.~Galley, C.~Brockett, J.~Gao, and B.~Dolan, ``A diversity-promoting
  objective function for neural conversation models,'' in \emph{Proceedings of
  the 2016 Conference of the North American Chapter of the Association for
  Computational Linguistics: Human Language Technologies}, 2016, pp. 110--119.

\bibitem[Serban et~al.(2017)Serban, Sordoni, Lowe, Charlin, Pineau, Courville,
  and Bengio]{serban2017hierarchical}
I.~V. Serban, A.~Sordoni, R.~Lowe, L.~Charlin, J.~Pineau, A.~Courville, and
  Y.~Bengio, ``A hierarchical latent variable encoder-decoder model for
  generating dialogues,'' in \emph{Thirty-First AAAI Conference on Artificial
  Intelligence}, 2017.

\bibitem[Zhao et~al.(2017)Zhao, Zhao, and Eskenazi]{zhao2017learning}
T.~Zhao, R.~Zhao, and M.~Eskenazi, ``Learning discourse-level diversity for
  neural dialog models using conditional variational autoencoders,'' in
  \emph{Proceedings of the 55th Annual Meeting of the Association for
  Computational Linguistics (Volume 1: Long Papers)}, 2017, pp. 654--664.

\bibitem[Cao and Clark(2017)]{cao2017latent}
K.~Cao and S.~Clark, ``Latent variable dialogue models and their diversity,''
  in \emph{Proceedings of the 15th Conference of the European Chapter of the
  Association for Computational Linguistics: Volume 2, Short Papers}, 2017, pp.
  182--187.

\bibitem[Ruan et~al.(2019)Ruan, Ling, Liu, Chen, and
  Indurkhya]{ruan2019condition}
Y.-P. Ruan, Z.-H. Ling, Q.~Liu, Z.~Chen, and N.~Indurkhya,
  ``Condition-transforming variational autoencoder for conversation response
  generation,'' in \emph{ICASSP 2019-2019 IEEE International Conference on
  Acoustics, Speech and Signal Processing (ICASSP)}.\hskip 1em plus 0.5em minus
  0.4em\relax IEEE, 2019, pp. 7215--7219.

\bibitem[Zhou et~al.(2018)Zhou, Huang, Zhang, Zhu, and Liu]{zhou2018emotional}
H.~Zhou, M.~Huang, T.~Zhang, X.~Zhu, and B.~Liu, ``Emotional chatting machine:
  Emotional conversation generation with internal and external memory,'' in
  \emph{Thirty-Second AAAI Conference on Artificial Intelligence}, 2018.

\bibitem[Song et~al.(2019)Song, Zheng, Liu, Xu, and Huang]{song2019generating}
Z.~Song, X.~Zheng, L.~Liu, M.~Xu, and X.-J. Huang, ``Generating responses with
  a specific emotion in dialog,'' in \emph{Proceedings of the 57th Annual
  Meeting of the Association for Computational Linguistics}, 2019, pp.
  3685--3695.

\bibitem[Zhou and Wang(2018)]{zhou2018mojitalk}
X.~Zhou and W.~Y. Wang, ``{MojiTalk}: Generating emotional responses at
  scale,'' in \emph{Proceedings of the 56th Annual Meeting of the Association
  for Computational Linguistics (Volume 1: Long Papers)}, 2018, pp. 1128--1137.

\bibitem[Radford et~al.(2018)Radford, Narasimhan, Salimans, and
  Sutskever]{radford2018improving}
\BIBentryALTinterwordspacing
A.~Radford, K.~Narasimhan, T.~Salimans, and I.~Sutskever, ``Improving language
  understanding by generative pre-training,'' \emph{OpenAI Blog}, 2018.
  [Online]. Available: \url{https://s3-us-west-2. amazonaws.
  com/openai-assets/researchcovers/languageunsupervised/language understanding
  paper. pdf}
\BIBentrySTDinterwordspacing

\bibitem[Devlin et~al.(2019)Devlin, Chang, Lee, and Toutanova]{devlin2018bert}
J.~Devlin, M.-W. Chang, K.~Lee, and K.~Toutanova, ``B{ERT}: Pre-training of
  deep bidirectional transformers for language understanding,'' in
  \emph{Proceedings of the 2019 Conference of the North American Chapter of the
  Association for Computational Linguistics: Human Language Technologies,
  Volume 1 (Long and Short Papers)}, 2019, pp. 4171--4186.

\bibitem[Yang et~al.(2019)Yang, Dai, Yang, Carbonell, Salakhutdinov, and
  Le]{yang2019xlnet}
Z.~Yang, Z.~Dai, Y.~Yang, J.~Carbonell, R.~R. Salakhutdinov, and Q.~V. Le,
  ``{XLNet}: Generalized autoregressive pretraining for language
  understanding,'' in \emph{Advances in neural information processing systems},
  2019, pp. 5754--5764.

\bibitem[Dong et~al.(2019)Dong, Yang, Wang, Wei, Liu, Wang, Gao, Zhou, and
  Hon]{dong2019unified}
L.~Dong, N.~Yang, W.~Wang, F.~Wei, X.~Liu, Y.~Wang, J.~Gao, M.~Zhou, and H.-W.
  Hon, ``Unified language model pre-training for natural language understanding
  and generation,'' in \emph{Advances in Neural Information Processing
  Systems}, 2019, pp. 13\,042--13\,054.

\bibitem[Elazar and Goldberg(2018)]{elazar2018adversarial}
Y.~Elazar and Y.~Goldberg, ``Adversarial removal of demographic attributes from
  text data,'' in \emph{Proceedings of the 2018 Conference on Empirical Methods
  in Natural Language Processing}, 2018, pp. 11--21.

\bibitem[Lample et~al.(2018)Lample, Subramanian, Smith, Denoyer, Ranzato, and
  Boureau]{lample2018multiple}
G.~Lample, S.~Subramanian, E.~Smith, L.~Denoyer, M.~Ranzato, and Y.-L. Boureau,
  ``Multiple-attribute text rewriting,'' in \emph{International Conference on
  Learning Representations}, 2018.

\bibitem[Kingma and Welling(2014)]{KingmaW13}
\BIBentryALTinterwordspacing
D.~P. Kingma and M.~Welling, ``Auto-encoding variational bayes,'' in \emph{2nd
  International Conference on Learning Representations, {ICLR} 2014, Banff, AB,
  Canada, April 14-16, 2014, Conference Track Proceedings}, 2014. [Online].
  Available: \url{http://arxiv.org/abs/1312.6114}
\BIBentrySTDinterwordspacing

\bibitem[Shang et~al.(2016)Shang, Sakai, Lu, Li, Higashinaka, and
  Miyao]{shang2016overview}
L.~Shang, T.~Sakai, Z.~Lu, H.~Li, R.~Higashinaka, and Y.~Miyao, ``Overview of
  the {NTCIR}-12 short text conversation task.'' in \emph{NTCIR}, 2016.

\bibitem[Maaten and Hinton(2008)]{maaten2008visualizing}
L.~v.~d. Maaten and G.~Hinton, ``Visualizing data using t-{SNE},''
  \emph{Journal of machine learning research}, vol.~9, no. Nov, pp. 2579--2605,
  2008.

\end{thebibliography}

%
\vspace{-4mm}
\begin{IEEEbiography}[{\includegraphics[width=1in,height=1.25in,clip,keepaspectratio]{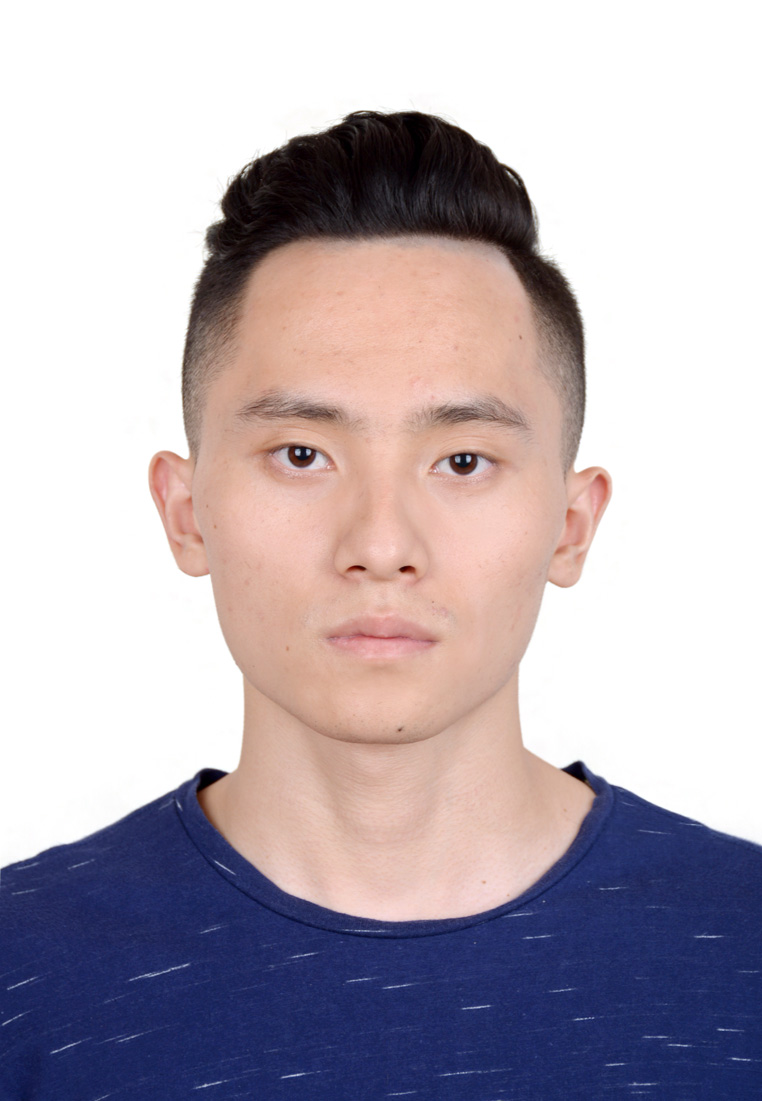}}]{Yu-Ping Ruan} received a B.S. degree from the School of Communication and Information Engineering, Shanghai University (SHU), in 2015 and received a Ph.D. degree in signal and information processing from the University of Science and Technology of China (USTC) in 2020. His research interests include natural language understanding, natural text generation and deep learning.
\end{IEEEbiography}

\vspace{-4mm}
\begin{IEEEbiography}[{\includegraphics[width=1in,height=1.25in,clip,keepaspectratio]{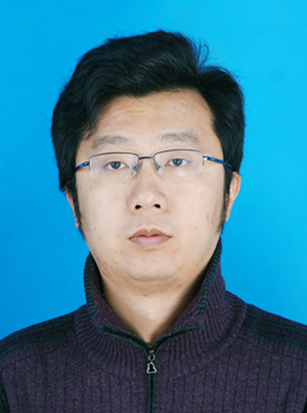}}]{Zhen-Hua Ling} (M'10 SM'19) received a B.E. degree in electronic information engineering and M.S. and Ph.D. degrees in signal and information processing from the University of Science and Technology of China, Hefei, China, in 2002, 2005, and 2008, respectively. From October 2007 to March 2008, he was a Marie Curie Fellow with the Centre for Speech Technology Research, University of Edinburgh, Edinburgh, U.K. From July 2008 to February 2011, he was a joint postdoctoral researcher with the University of Science and Technology of China and iFLYTEK Co. Ltd., Hefei, China. He is currently an Associate Professor with the University of Science and Technology of China. He also worked at the University of Washington, Seattle, WA, USA, as a Visiting Scholar from August 2012 to August 2013. His research interests include speech processing, speech synthesis, voice conversion, and natural language processing. He was the recipient of the IEEE Signal Processing Society Young Author Best Paper Award in 2010. He was an Associate Editor of IEEE/ACM Transactions on Audio, Speech, and Language Processing from 2014 to 2018.
\end{IEEEbiography}




\end{document}